\documentclass{article}

\usepackage{arxiv}
\usepackage[utf8]{inputenc}
\usepackage[T1]{fontenc}
\usepackage{hyperref}
\usepackage{url}
\usepackage{booktabs}
\usepackage{amsfonts}
\usepackage{nicefrac}
\usepackage{microtype}
\usepackage{graphicx}
\usepackage{booktabs} 
\usepackage{multirow} 
\usepackage{makecell} 
\usepackage{natbib}
\usepackage{threeparttable}
\usepackage{float}
\usepackage{amsmath}
\usepackage{tcolorbox}
\graphicspath{{./images/}}

\title{Masked‑and‑Reordered Self‑Supervision for Reinforcement Learning from Verifiable Rewards}

\author{
  Zhen Wang\thanks{Corresponding author. \texttt{wangz@dp.tech}} \\
  DP Technology\\
  \And
  Zhifeng Gao \\
  DP Technology\\
  \And
  Guolin Ke\\
  DP Technology\\
}

\begin{document}

\renewcommand{\thefootnote}{\fnsymbol{footnote}}
\maketitle
\setcounter{footnote}{0}
\renewcommand{\thefootnote}{\arabic{footnote}}

\begin{abstract}
Test-time scaling has been shown to substantially improve large language models’ (LLMs) mathematical reasoning. However, for a large portion of mathematical corpora, especially theorem proving, RLVR’s scalability is limited: intermediate reasoning is crucial, while final answers are difficult to directly and reliably verify. Meanwhile, token-level SFT often degenerates into rote memorization rather than inducing longer chains of thought. Inspired by BERT’s self-supervised tasks, we propose MR-RLVR (Masked-and-Reordered RLVR), which constructs process-level self-supervised rewards via “masked-then-fill” and “step reordering” to extract learnable signals from intermediate reasoning. Our training pipeline comprises two stages: we first perform self-supervised training on sampled mathematical calculation and proof data; we then conduct RLVR fine-tuning on mathematical calculation datasets where only outcomes are verifiable. We implement MR-RLVR on Qwen2.5-3B and DeepSeek-R1-Distill-Qwen-1.5B, and evaluate on AIME24, AIME25, AMC23, and MATH500. Under a fixed sampling and decoding budget, MR-RLVR achieves average relative gains over the original RLVR of +9.86\% Pass@1, +5.27\% Pass@5, and +4.00\% Pass@8. These results indicate that incorporating process-aware self-supervised signals can effectively enhance RLVR’s scalability and performance in only outcome-verifiable settings.
\end{abstract}

\section{Introduction}

Large language models (LLMs) have recently made rapid progress on mathematical and scientific reasoning tasks, driven by techniques such as chain-of-thought prompting, diverse sampling, and test-time scaling. Reinforcement learning (RL) has further improved performance on reasoning-intensive tasks including mathematical problem solving, code generation, and program synthesis~\citep{ yang2025qwen3technicalreport, guo2025deepseek}. A central question in these settings is how to design reward signals that guide models toward reliable and generalizable reasoning strategies. Verifiable rewards obtained by programmatically checking final answers or executing unit tests offer a practical solution: whether a model output satisfies predefined symbolic or numeric constraints can often be determined automatically, providing scalable and low-cost supervision for RL~\citep{guo2024deepseekcoderlargelanguagemodel, yang2024qwen25mathtechnicalreportmathematical}. Reinforcement Learning from Verifiable Rewards (RLVR)~\citep{shao2024deepseekmath} instantiates this idea by directly optimizing policies to pass symbolic or numerical checks at the level of final answers, and has shown strong performance on code generation and mathematical reasoning tasks~\citep{ shao2024deepseekmath, guo2025deepseek}.

However, terminally verifiable rewards primarily constrain the final answer. To further improve complex multi-step reasoning, it is crucial yet challenging to construct equally informative training signals for the \emph{intermediate} reasoning process. One line of work explicitly leverages intermediate steps: process supervision and Process Reward Model (PRM) frameworks~\citep{lightman2023letsverifystepstep, guan2025rstarmathsmallllmsmaster} provide step-level supervision by scoring or classifying intermediate steps, improving stability and interpretability in mathematical reasoning. These approaches, however, typically require large-scale, high-quality human annotations, and in complex theorem-proving scenarios it is inherently difficult to decide whether a local step is reasonable. This leads to high annotation costs and limits scalability to diverse large-scale corpora. In addition, token-level supervised fine-tuning (SFT) on chain-of-thought data often degenerates into imitating specific solution templates rather than learning transferable reasoning strategies~\citep{lightman2023letsverifystepstep}. Human-centric supervision thus struggles to simultaneously achieve low annotation cost, scalability, and transferability of reasoning ability.

RLVR removes the need for step-level labels but provides only weak constraints on the intermediate reasoning trajectory. As a result, it is susceptible to \emph{process hallucinations}: the model may generate plausible-looking yet incorrect or redundant reasoning steps, and such errors are often difficult to detect and correct using only terminal verification. In many mathematical and theorem-proving datasets, correctness at the step level is important in its own right, yet hard to verify via a unified, low-cost programmatic procedure. Complementary to process supervision, recent work has explored self-supervised signals or existing trajectories as denser rewards for reasoning, for example by using model confidence in reference answers as intrinsic rewards or by designing next-segment reasoning objectives from expert traces and pretraining corpora and converting them into RL training signals~\citep{yu2025rlprextrapolatingrlvrgeneral, li2025reinforcementlearningpretrainingdata, deng2025supervisedreinforcementlearningexpert}. These approaches mitigate reward sparsity without additional human annotations or domain-specific verifiers, suggesting that combining self-supervision with RL is a promising direction for strengthening multi-step reasoning.

We ask whether it is possible to extract \emph{process-level} signals directly from existing mathematical reasoning trajectories and convert them into verifiable rewards that are compatible with RLVR, without relying on additional human process annotations or explicit expert action sequences. Our starting point is the observation that self-supervised destruction--reconstruction objectives, such as those used in BERT\citep{devlin2019bertpretrainingdeepbidirectional, raffel2023exploringlimitstransferlearning}, help models capture semantic and structural dependencies within context by masking and reconstructing missing spans. This property naturally aligns with modeling the constraints and dependencies between steps in multi-step reasoning.

In this work, we propose MR-RLVR (Masked-and-Reordered RLVR), which augments RLVR with dense, structured process-level self-supervision derived from mathematical reasoning trajectories. Under a setting where only terminal rewards are externally verifiable, we construct internal process rewards that can be computed automatically and integrated into RL training. Concretely, we design two process-aware tasks on proof-style and computation-style trajectories: (1) Masked-Then-Fill, which masks key formulas, reasoning steps, or theorem invocations and requires the model to reconstruct the missing content given surrounding context; and (2) Step Reordering, which shuffles reasoning steps and asks the model to recover a coherent logical order. For both tasks, we define process-level rewards based on the match between generated spans and reference reasoning at the levels of mathematical entities and text, and on the agreement between predicted and reference step orders. These rewards can be computed automatically from existing trajectories and used directly as RL signals, without any additional human annotation.

MR-RLVR adopts a simple two-stage training framework. In Stage I, we use only the process-level rewards described above to perform RLVR updates on mathematical corpora that jointly cover proof-style and computation-style reasoning, encouraging the policy to produce reasoning processes with more coherent local logic and clearer step dependencies. In Stage II, starting from the Stage I checkpoint, we fine-tune the model on computational math problems with programmatically verifiable final answers, using only terminally verifiable rewards. The model first learns better reasoning structure under dense process-level signals, and then adapts to verifiable tasks under sparse but precise terminal supervision. This combination aims to improve the stability and scalability of RLVR on complex mathematical reasoning tasks without increasing human annotation cost, and to reduce process hallucinations.

Our contributions are summarized as follows:
\begin{enumerate}
    \item \textbf{Process-level self-supervision as verifiable rewards.} We propose a framework that designs Mask-Then-Fill and Step Reordering tasks on mathematical reasoning trajectories and converts their outcomes into process-level rewards based on mathematical-entity matching and ordering consistency. This allows RLVR to receive fine-grained process supervision without any human process annotations.
    \item \textbf{Two-stage MR-RLVR training under terminal-verifiable supervision.} We introduce a two-stage training procedure that first performs RLVR pretraining with process-level rewards on diverse proof and computational reasoning corpora, and then applies outcome-level RLVR fine-tuning on computational problems with verifiable final answers, alleviating exploration difficulties under sparse rewards.
    \item \textbf{Empirical gains and data efficiency on mathematical reasoning benchmarks.} On Qwen2.5-3B and DeepSeek-R1-Distill-Qwen-1.5B, MR-RLVR consistently outperforms a GRPO baseline on AIME24, AIME25, AMC23, and MATH500, achieving an average relative improvement of about \(9.86\%\) in \(\text{Pass@1}\), \(5.27\%\) in \(\text{Pass@5}\) and \(4.00\%\) in \(\text{Pass@8}\) under a fixed sampling budget. In low-data regimes, MR-RLVR also exhibits better sample efficiency than standard RLVR.
\end{enumerate}

\section{Related work}
\paragraph{PRM and RLVR for math reasoning.}
From the perspective of training signals, large-model training for mathematical reasoning typically follows two lines: \emph{process supervision} and \emph{outcome-verifiable rewards}. The former is exemplified by Process Reward Models (PRMs) combined with MCTS-style search, which score intermediate steps and use step-level value estimates to guide tree expansion and pruning, thereby achieving stable, interpretable, and search-capable reinforcement reasoning on math and code tasks~\citep{lightman2023letsverifystepstep, wang2024mathshepherdverifyreinforcellms, zhang2024restmctsllmselftrainingprocess, guan2025rstarmathsmallllmsmaster}. However, such methods require expensive step-level annotations (or large-model scoring) as well as additional search infrastructure. In contrast, the RLVR line avoids explicit step labels and relies solely on programmable outcome verifiers that score final answers or executable code~\citep{shao2024deepseekmath}. On automatically gradable benchmarks such as AIME, AMC, and MATH, systems including DeepSeek-Math, and Qwen2.5-Math have demonstrated substantial gains in mathematical and code reasoning under this paradigm~\citep{guo2024deepseekcoderlargelanguagemodel, shao2024deepseekmath, guo2025deepseek, yang2024qwen25mathtechnicalreportmathematical, yang2025qwen3technicalreport}. Nevertheless, standard RLVR imposes almost no constraints on the intermediate reasoning process and is thus prone to process hallucinations and redundant steps. In comparison, MR-RLVR preserves the RLVR assumption of relying only on outcome-verifiable rewards, without introducing external PRMs or human step labels; instead, it automatically constructs self-supervised tasks such as masked reconstruction and step permutation on existing mathematical reasoning trajectories, converts their completion quality into process-level rewards usable by RL, and combines them with a second-stage RLVR training based on outcome rewards.

\paragraph{Self-supervised process signals for reasoning tasks.}
A complementary line of work explores self-supervised \emph{process signals} for reasoning tasks. ClozeMath adapts text-infilling and PrefixLM objectives to the mathematical setting by masking intermediate equations during supervised fine-tuning and requiring the model to recover them, thereby strengthening the modeling of key mathematical entities and local structure; however, its self-supervised signal is only used as an SFT loss and is not explicitly converted into RL rewards~\citep{pham2025clozemathimprovingmathematicalreasoning}. RLPR, RLPT, and SRL instead embed self-supervised signals into reinforcement learning or preference-optimization frameworks: RLPR uses the model’s (relative) generation probability of a reference answer as an intrinsic reward, extending RLVR to general domains without human scoring~\citep{yu2025rlprextrapolatingrlvrgeneral}; RLPT defines a next-chunk prediction objective on large-scale unlabeled text and employs an auxiliary model to score semantic consistency between the prediction and the ground-truth continuation as a reward~\citep{li2025reinforcementlearningpretrainingdata}; SRL decomposes expert solutions into sequences of actions and uses action-level similarity as rewards to guide stepwise imitation of expert trajectories~\citep{deng2025supervisedreinforcementlearningexpert}. These works demonstrate that probabilities, semantic consistency, and action similarity can all serve as effective intrinsic rewards, but their reward designs are mostly centered on the consistency or similarity of whole answers or relatively long segments, with limited specialization for key entities, local logical dependencies, and step ordering in mathematical reasoning. MR-RLVR instead directly designs fine-grained, structure-aware self-supervised tasks (masked refilling and step permutation) on mathematical reasoning trajectories and converts task outcomes into automatically computable process-level rewards that integrate seamlessly into an RLVR pipeline.

\section{Preliminaries}
A large language model (LLM) can be regarded as a conditional probabilistic model \(\pi_\theta(y,z\mid x)\),  
where \(x\in\mathcal{X}\) denotes the input problem or context, \(z=(z_1,\dots,z_T)\) represents the reasoning trajectory,  
and \(y\in\mathcal{Y}\) denotes the final output.  
When a ground‑truth answer is available, it is denoted by \(y^\star\).  
The learning signal is provided by a \textit{verifiable reward} function \(r(x,z,y)\!\in\![0,1]\),  
which quantifies the degree to which a model‑generated solution satisfies the task specification.  
This reward is computed automatically through a programmatic evaluation that checks the logical or factual consistency of the model output.  
Depending on the task, \(r\) may be instantiated through numerical tolerance scoring, symbolic or textual equivalence, 
structured output validation, or code‑level unit and integration tests.   

\paragraph{RLVR objective.}
Reinforcement Learning from Verifiable Rewards (RLVR) maximizes the expected verifiable reward while penalizing divergence from a reference policy through a Kullback–Leibler (KL) regularization term.  
Formally, the optimization objective is
\begin{equation}
\max_{\theta}\;
\mathbb{E}_{x\sim\mathcal{D},\,(z,y)\sim\pi_\theta(z,y\mid x)} 
\!\left[ r(x,z,y) \right]
\;-\;
\beta
\,\mathbb{E}_{x\sim\mathcal{D}}\!\left[
  \mathrm{KL}\!\left(\pi_\theta(z,y\mid x)\;\|\;\pi_{\text{ref}}(z,y\mid x)\right)
\right],
\end{equation}
where \(\pi_{\text{ref}}\) is a fixed or exponentially moving‑averaged (EMA) reference policy,  
and \(\beta>0\) controls the trade‑off between maximizing the verifiable reward and maintaining proximity to the reference distribution.  
The first term encourages the model to generate reasoning paths and answers that satisfy programmatically verifiable conditions, whereas the KL penalty mitigates excessive policy drift and stabilizes training under the restricted data regime typical of verifiable tasks.  
This formulation parallels the general form of Reward‑Regularized Policy Optimization (RRPO) and serves as the foundation for the GRPO update rule introduced below.

\paragraph{GRPO formulation.}
Generalized Reinforcement Policy Optimization (GRPO)\citep{ shao2024deepseekmath}optimizes the policy by maximizing a clipped surrogate objective over groups of sampled outputs, providing stable updates within the RLVR framework.
For each input \(q\), a set of responses \(\{o_i\}_{i=1}^{G}\) is drawn from the old policy \(\pi_{\theta_{\mathrm{old}}}\),  
and the policy parameters are optimized under the objective
\begin{equation}
J_{\text{GRPO}}(\theta)
=\mathbb{E}\!\left[
\frac{1}{G}\sum_{i=1}^{G}\frac{1}{|o_i|}
\sum_{t=1}^{|o_i|}
\min\!\left(
\rho_{\theta,i,t}\hat{A}_{i,t},\;
\mathrm{clip}\!\left(\rho_{\theta,i,t},\,1-\epsilon,\,1+\epsilon\right)\hat{A}_{i,t}
\right)
\right]
-\beta\,D_{\mathrm{KL}}\!\left(\pi_\theta\,\|\,\pi_{\mathrm{ref}}\right),
\end{equation}
where \(\rho_{\theta,i,t}=\frac{\pi_\theta(o_{i,t}\mid q,o_{i,<t})}{\pi_{\theta_{\mathrm{old}}}(o_{i,t}\mid q,o_{i,<t})}\)  
is the per‑token likelihood ratio and \(\hat{A}_{i,t}\) is the corresponding advantage estimate.  
Outcome‑level supervision assigns each sequence \(o_i\) a normalized scalar reward  
\(\hat{A}_{i,t} = (r_i - \operatorname{mean}(r))/\operatorname{std}(r)\),  
shared across all tokens in that sequence.  
The KL regularization term constrains deviation from reference model, and is estimated using the per‑token unbiased form  
\begin{equation}
D_{\mathrm{KL}}\!\left(\pi_\theta\,\|\,\pi_{\mathrm{ref}}\right)
=
\frac{\pi_{\mathrm{ref}}(o_{i,t}\mid q,o_{i,<t})}{\pi_\theta(o_{i,t}\mid q,o_{i,<t})}
-\log\frac{\pi_{\mathrm{ref}}(o_{i,t}\mid q,o_{i,<t})}{\pi_\theta(o_{i,t}\mid q,o_{i,<t})}
-1,
\end{equation}
following the unbiased estimator proposed by~\citet{schulman2015high}. 
This formulation retains the stability properties of PPO while optimizing directly toward verifiable reward signals.

\section{Methodology}
\begin{figure}[htbp]
    \centering
    \includegraphics[trim=16mm 16mm 16mm 16mm, clip, width=\linewidth]{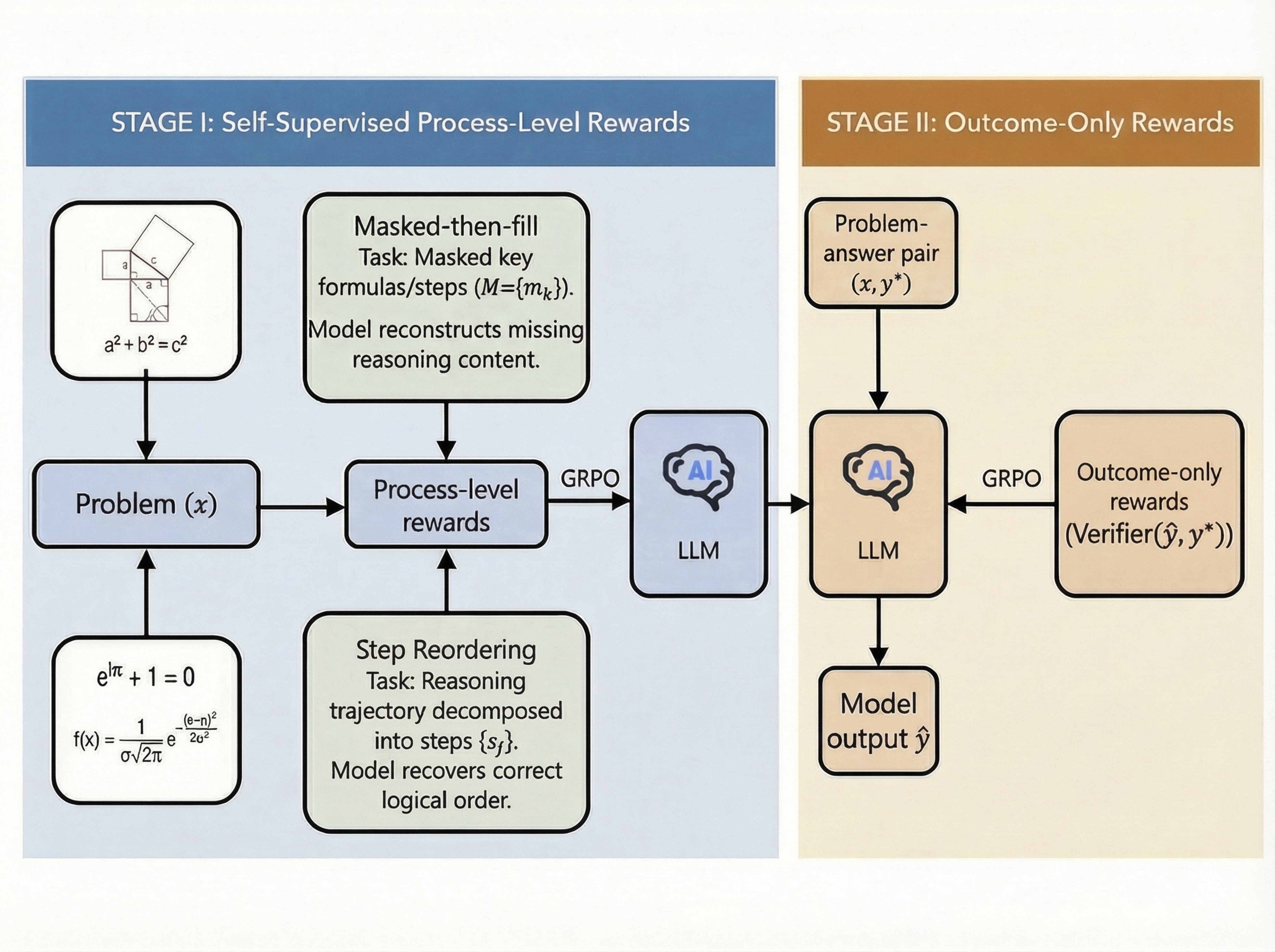}
    \vspace{-40pt}
    \caption{Overview of the MR-RLVR two-stage training framework. Stage I uses process-level rewards from self-supervised tasks; Stage II uses final-outcome rewards.}
    \label{fig:framework}
\end{figure}

MR‑RLVR enables models to exploit intermediate reasoning even when only final answers are verifiable. Training proceeds in two stages. In Stage I, we derive process‑level rewards from reasoning traces via process‑aware self‑supervised tasks (Masked‑Then‑Fill and Step Reordering) and run an initial RLVR phase using only these process‑level rewards. In Stage II, we fine‑tune with RLVR supervised exclusively by programmatically verifiable final‑answer rewards. This two‑stage design first shapes the policy distribution with process‑level signals and then trains under sparse outcome supervision, yielding more informative gradients and more stable exploration. Figure~\ref{fig:framework} illustrates the framework: Stage I uses only process‑level rewards; Stage II uses only final‑answer rewards.
\subsection{Process-aware self-supervised data curation}

In the first stage, we construct training data for process-aware self-supervised tasks, designed to guide the model in learning local logical consistency and step dependencies within reasoning trajectories. Specifically, we sample mathematical proof and computational reasoning problems to form the process-level self-supervised training corpus, while only the computational reasoning subset is used in Stage~II fine-tuning.  

For each problem $x$ and its corresponding reasoning trajectory and answer $(z, y)$, we not only filter semantically well-structured and symbolically valid reasoning texts, but also restructure overly verbose or redundant reasoning chains to obtain concise and logically coherent reasoning processes. Two structural transformations are then applied:

\paragraph{(1)Masked-Then-Fill task.}  
Key formulas, inference steps, or theorem invocations in the reasoning text are masked to create a list of masked positions $\mathcal{M}=\{m_k\}$, along with the corresponding ground-truth completions. This task requires the model to reconstruct missing reasoning content given contextual information.  
    
\paragraph{(2)Step Reordering task.}  
The reasoning trajectory is decomposed into ordered steps $\{s_j\}$, which are randomly permuted to form a perturbed sequence. The model is required to recover the correct order based on logical coherence.

After these operations, we denote the processed reasoning sequence as $\hat{z}$. The model input consists of the problem statement $x$ and the modified reasoning $\hat{z}$, while the supervision signal involves either token restoration or step order prediction.

\subsection{Process-level reward design for self-supervision}
During the self-supervised phase, the model constructs a process-level reward signal based on either the Masked-Then-Fill or Step Reordering task (only one of them is used per training run).  

\paragraph{(1) Masked-Then-Fill reward.}  
For masked samples, $h$ masked locations are randomly selected for evaluation. The reward is defined as the mean semantic match score between model completions and ground truths, measured by MathRuler for mathematical entity alignment and supplemented by textual similarity as fallback:  
\begin{equation}  
r_{\text{mask}}(x,\hat{z},y)  
= \frac{1}{h}\sum_{k=1}^h   
   \mathrm{Match}_{\text{entity}}\!\left(\hat{m}_k, m_k^\star\right).  
\end{equation}

\paragraph{(2) Step Reordering reward.}
For the step reordering task, we measure how well the model recovers the correct position of each reasoning step in the sequence.
Let \(o_{\text{true}}\) denote the reference order over \(n\) reasoning steps, and \(o_{\text{pred}}\) the permutation predicted by the model.
We denote by \(\mathrm{pos}_{\text{true}}(k)\) the index of step \(k\) in the reference order \(o_{\text{true}}\), and by
\(\mathrm{pos}_{\text{pred}}(k)\) its index in the predicted order \(o_{\text{pred}}\).
We define a normalized position-based distance
\[
d_{\text{pos}}(o_{\text{pred}}, o_{\text{true}})
= \frac{1}{n} \sum_{k=1}^{n}
\mathbb{I}\bigl[\mathrm{pos}_{\text{pred}}(k) \neq \mathrm{pos}_{\text{true}}(k)\bigr],
\]
where \(\mathbb{I}[\cdot]\) is the indicator function.
This distance \(d_{\text{pos}} \in [0,1]\) measures the fraction of steps that are placed at incorrect positions.

The corresponding step-order reward is then defined as
\[
r_{\text{order}}(x,\tilde{z},y)
= 1 - d_{\text{pos}}(o_{\text{pred}}, o_{\text{true}}),
\]
so that perfectly ordered sequences receive reward \(1\), while sequences in which all steps are misplaced receive reward \(0\).
This reward naturally lies in \([0,1]\) and can be directly combined with the masked-then-fill reward within the MR-RLVR framework.

Accordingly, the process-level reward is defined as
\begin{equation}
r_{\text{proc}}(x,\tilde{z},y) = 
\mathbb{I}_{\text{mask}} \cdot r_{\text{mask}}(x,\tilde{z},y) 
+ \mathbb{I}_{\text{order}} \cdot r_{\text{order}}(x,\tilde{z},y),
\end{equation}
where $\mathbb{I}_{\text{mask}}$, $\mathbb{I}_{\text{order}} \in \{0,1\}$ are indicator functions for the Masked-Then-Fill and Step Reordering tasks with $\mathbb{I}_{\text{mask}} + \mathbb{I}_{\text{order}} = 1$.

\subsection{Stage II: fine-tuning with outcome-only rewards}
In the second stage, we initialize from the checkpoint obtained through process-level reinforcement learning and fine-tune on programmatically verifiable problem instances. The model now generates \emph{complete} reasoning trajectories from the problem statement $x$
and the model generates both reasoning $z$ and answer $y$, receiving supervision from the final-outcome reward $r_{\text{final}}(y^\star, y)$.  

Only computational reasoning tasks are included here, as they feature open-ended reasoning trajectories but deterministic final answers. During fine-tuning, the model receives a binary verifiable reward:  
\begin{equation}  
r_{\text{final}}(y^\star, y) = \mathbb{I}\bigl[\text{Verify}(y^\star, y) = \text{True}\bigr],  
\end{equation}  
where $y^\star$ is the ground-truth answer and $\text{Verify}(y^\star, y)$ denotes symbolic and numerical verification comparing the generated answer $y$ against $y^\star$.  

This signal is used to perform RLVR optimization under the GRPO objective. The two-stage scheme first shapes the reasoning distribution using dense process-level supervision, then refines it with sparse but precise outcome rewards, yielding stable optimization and verifiably correct multi-step reasoning behavior.    

\section{Experiment}

\subsection{Experimental Setup}
\paragraph{Training Data Curation} To accommodate diverse reasoning styles, we construct our training corpus from two data sources: DeepTheorem\citep{zhang2025deeptheoremadvancingllmreasoning} primarily contains theorem-proving problems, while DeepMath \citep{he2025deepmath103klargescalechallengingdecontaminated} focuses on computational reasoning tasks. We sample 10k problems equally from both datasets. Since the reasoning traces in DeepMath are generated by DeepSeek models and contain multiple internal verification steps that may lead to information leakage within trajectories, we refine all DeepMath samples using \texttt{GPT-o3-mini} to ensure clean, step-by-step reasoning without redundant self-verification. As identifying key theorems, formulas, and reasoning steps for masking or reordering is non-trivial, we employ \texttt{DeepSeek-R1-0528} to process the refined reasoning trajectories and generate task-specific annotations: masked positions for the Masked-Then-Fill task and step boundaries for the Step Reordering task (prompt templates in Appendix~\ref{app:prompts}). To avoid trivial masking tasks, we retain only samples with at least 7 masked positions. From the filtered pool, we select 10k samples for each of the two tasks (Masked-Then-Fill and Step Reordering), randomly sampling 20k instances for Stage~I training and 6k for validation. For Stage~II fine-tuning, we select 5k computational reasoning samples from the Stage~I training data as the training set, with 1.5k held out for validation, ensuring that Stage~II focuses on verifiable computational problems with deterministic answers.  

\paragraph{Model Configuration} 
We conduct experiments on Qwen2.5-3B-Base\citep{qwen2025qwen25technicalreport} and DeepSeek-R1-Distill-Qwen-1.5B\citep{guo2025deepseek}.  We employ the GRPO objective for reinforcement learning. All experiments are conducted within the verl framework \citep{Sheng_2025} on a single node with 8 NVIDIA A100 (80 GB) or A800 (80 GB) GPUs. To optimize GPU memory usage, several parameters differ slightly between Qwen-3B and DeepSeek-R1-Distill-Qwen-1.5. Detailed experimental hyperparameters can be found in the Appendix~\ref{app:impl}.

\paragraph{Evaluation Setup} 
We evaluate reasoning performance on four challenging mathematical benchmarks: AIME 2024, AIME 2025, AMC 2023 \citealp{li2024numinamath}, and MATH500 \citealp{hendrycks2021measuring}. We report the unbiased estimator of Pass@$k$~\cite{chen2021evaluatinglargelanguagemodels}, defined as 
\begin{equation}  
\text{Pass@}k = \mathbb{E}_{x \sim \mathcal{D}} \left[ 1 - \frac{\binom{n-c}{k}}{\binom{n}{k}} \right],  
\end{equation}  
where $n$ is the number of generated solutions per problem, $c$ is the number of correct solutions, and $k \in \{1, 5, 8\}$ denotes the number of attempts allowed. We set $n = 64$ for all evaluations. During inference, we use nucleus sampling with temperature 0.6, top-$p$ 0.95, and a maximum generation length of 4096 tokens. Answers are verified programmatically through symbolic computation using MathRuler~\cite{mathruler} and text matching.   

\subsection{Main Results}

\begin{table*}[htbp]  
\centering  
\caption{Performance comparison across mathematical reasoning benchmarks.}  
\label{tab:main_results}  
\footnotesize  
\setlength{\tabcolsep}{3.5pt}  
\begin{tabular}{@{}lcccccccccc@{}}  
\toprule  
\multicolumn{10}{c}{\textbf{Qwen2.5-3B}} \\
\midrule  
\multirow{2}{*}{\textbf{Benchmark}} & \multicolumn{3}{c}{\textbf{Pass@1 (\%)}} & \multicolumn{3}{c}{\textbf{Pass@5 (\%)}} & \multicolumn{3}{c}{\textbf{Pass@8 (\%)}} \\
\cmidrule(lr){2-4} \cmidrule(lr){5-7} \cmidrule(lr){8-10}  
& Base & +GRPO & +MR-RLVR & Base & +GRPO & +MR-RLVR & Base & +GRPO & +MR-RLVR \\
\midrule  
AIME24 & 1.93 & 5.63 & \textbf{6.30} {\scriptsize ($\uparrow$12.04\%)} & 7.48 & 14.29 & 13.20 {\scriptsize ($\downarrow$7.61\%)} & 10.23 & 17.29 & 15.43 {\scriptsize ($\downarrow$10.74\%)} \\
AIME25 & 0.73 & 2.03 & \textbf{2.76} {\scriptsize ($\uparrow$35.98\%)} & 3.58 & 8.53 & \textbf{10.44} {\scriptsize ($\uparrow$22.44\%)} & 5.13 & 12.10 & \textbf{14.05} {\scriptsize ($\uparrow$16.11\%)} \\
AMC23 & 14.06 & 36.13 & \textbf{40.82} {\scriptsize ($\uparrow$12.98\%)} & 41.70 & 60.39 & \textbf{64.48} {\scriptsize ($\uparrow$6.82\%)} & 50.89 & 66.29 & \textbf{69.80} {\scriptsize ($\uparrow$5.29\%)} \\
MATH500 & 27.75 & 63.30 & \textbf{65.87} {\scriptsize ($\uparrow$4.06\%)} & 62.97 & 79.83 & \textbf{80.94} {\scriptsize ($\uparrow$1.39\%)} & 70.97 & 83.29 & \textbf{83.85} {\scriptsize ($\uparrow$0.67\%)} \\
\midrule  
\midrule  
\multicolumn{10}{c}{\textbf{DeepSeek-R1-Distill-Qwen-1.5B}} \\
\midrule  
\multirow{2}{*}{\textbf{Benchmark}} & \multicolumn{3}{c}{\textbf{Pass@1 (\%)}} & \multicolumn{3}{c}{\textbf{Pass@5 (\%)}} & \multicolumn{3}{c}{\textbf{Pass@8 (\%)}} \\
\cmidrule(lr){2-4} \cmidrule(lr){5-7} \cmidrule(lr){8-10}  
& Base & +GRPO & +MR-RLVR & Base & +GRPO & +MR-RLVR & Base & +GRPO & +MR-RLVR \\
\midrule  
AIME24 & 9.17 & 18.70 & \textbf{19.43} {\scriptsize ($\uparrow$3.90\%)} & 21.93 & 36.40 & \textbf{36.96} {\scriptsize ($\uparrow$1.54\%)} & 26.00 & 41.98 & \textbf{42.08} {\scriptsize ($\uparrow$0.24\%)} \\
AIME25 & 10.62 & 15.94 & \textbf{17.24} {\scriptsize ($\uparrow$8.17\%)} & 23.10 & 27.40 & \textbf{31.72} {\scriptsize ($\uparrow$15.77\%)} & 26.43 & 29.50 & \textbf{35.43} {\scriptsize ($\uparrow$20.12\%)} \\
AMC23 & 36.84 & 62.30 & \textbf{63.01} {\scriptsize ($\uparrow$1.14\%)} & 66.57 & 84.23 & \textbf{85.62} {\scriptsize ($\uparrow$1.65\%)} & 72.98 & 89.30 & \textbf{89.48} {\scriptsize ($\uparrow$0.20\%)} \\
MATH500 & 60.80 & 78.05 & \textbf{78.51} {\scriptsize ($\uparrow$0.59\%)} & 85.04 & 90.08 & \textbf{90.25} {\scriptsize ($\uparrow$0.19\%)} & 88.38 & 91.85 & \textbf{91.97} {\scriptsize ($\uparrow$0.13\%)} \\
\bottomrule  
\end{tabular}  
\begin{tablenotes}  
\small  
\item We report Pass@$k$ ($k \in \{1, 5, 8\}$) with $n=64$ samples per problem. Arrows indicate relative improvement (\%) over the GRPO baseline. \textbf{Bold} indicates the best performance within each model family.  
\end{tablenotes}  
\end{table*}  

Table~\ref{tab:main_results} presents the main experimental results comparing our MR-RLVR method against the GRPO baseline across four mathematical reasoning benchmarks. Overall, MR-RLVR demonstrates consistent performance improvements over GRPO, particularly in challenging scenarios where baseline performance is relatively low. On the Qwen2.5-3B model, we observe substantial gains on AIME 2025, with Pass@1, Pass@5, and Pass@8 achieving relative improvements of 35.98\%, 22.44\%, and 16.11\%, respectively. Similar trends are evident in AIME24 Pass@1 (+12.04\%) and across all metrics on AMC23 (+5.29\% to +12.98\%).

Interestingly, the performance gains exhibit interesting patterns across different model architectures and task difficulties. In terms of average improvement magnitude, MR-RLVR yields significantly larger gains on Qwen2.5-3B (average 8.29\%) compared to DeepSeek-R1-Distill-Qwen-1.5B (average 4.47\%). This discrepancy may be attributed to the fact that the former has only undergone basic pretraining without complex post-training procedures, thus offering greater room for optimization through verification enhancement. Another notable observation is that both models show relatively limited improvements on MATH500 (0.13\%-4.06\%). This is not only because the baseline performance is already high (Pass@1>78\%), but more importantly, the problems in MATH500 are relatively simple and the models have largely mastered their solution patterns, resulting in minimal marginal gains from multi-round verification. In contrast, our method demonstrates much stronger value on competition-level problems such as AIME and AMC. We also observe minor performance degradation on AIME24 Pass@5/Pass@8 for Qwen2.5-3B, which we attribute to the inherent variance in the sampling process with n=64 samples.

These results validate the effectiveness of our MR-RLVR framework in improving mathematical reasoning capabilities. The consistent gains on challenging benchmarks (AIME24, AIME25, AMC23) demonstrate that MR-RLVR successfully enhances the model's ability to generate and verify correct solutions, particularly in scenarios where baseline performance leaves substantial room for improvement. Meanwhile, the more modest improvements on relatively simple tasks like MATH500 indicate that our method provides maximum value when applied to high-difficulty problems at the boundary of model competence, which aligns well with the design principle of MR-RLVR, to tackle complex reasoning tasks not yet fully mastered by the model through iterative refinement.

\subsection{More Analysis about MR-RLVR}
\subsubsection{Data Efficiency Analysis}
\begin{table*}[htbp]  
\centering  
\caption{Performance comparison of DeepSeek-R1-Distill-Qwen-1.5B under different training data scales. }  
\label{tab:data_efficiency}  
\footnotesize  
\setlength{\tabcolsep}{3pt}  
\begin{tabular}{@{}lcccccccccccc@{}}  
\toprule  
\multirow{2}{*}{\textbf{Method}} & \multicolumn{3}{c}{\textbf{AIME24}} & \multicolumn{3}{c}{\textbf{AIME25}} & \multicolumn{3}{c}{\textbf{AMC23}} & \multicolumn{3}{c}{\textbf{MATH500}} \\
\cmidrule(lr){2-4} \cmidrule(lr){5-7} \cmidrule(lr){8-10} \cmidrule(lr){11-13}  
& P@1 & P@5 & P@8 & P@1 & P@5 & P@8 & P@1 & P@5 & P@8 & P@1 & P@5 & P@8 \\
\midrule  
Base Model & 9.17 & 21.93 & 26.00 & 10.62 & 23.10 & 26.43 & 36.84 & 66.57 & 72.98 & 60.80 & 85.04 & 88.38 \\
\midrule  
\multicolumn{13}{c}{\textit{Training with 1k samples}} \\
\midrule  
+GRPO & 11.09 & 24.62 & 28.69 & \textbf{11.98} & 24.56 & 28.02 & 41.99 & 70.81 & 77.11 & 65.24 & 86.28 & 89.01 \\
+MR-RLVR & \textbf{11.09} & \textbf{26.90} & \textbf{31.84} & 11.93 & \textbf{25.45} & \textbf{28.91} & \textbf{42.73} & \textbf{72.82} & \textbf{79.25} & \textbf{65.67} & \textbf{86.49} & \textbf{89.24} \\
\midrule  
\multicolumn{13}{c}{\textit{Training with 3k samples}} \\
\midrule  
+GRPO & \textbf{16.67} & 32.41 & 37.22 & 15.31 & 28.67 & 31.97 & 58.01 & 81.86 & \textbf{86.91} & 76.40 & \textbf{89.77} & \textbf{91.79}  \\
+MR-RLVR & 16.56 & \textbf{35.13} & \textbf{40.97} & \textbf{15.57} & \textbf{29.37} & \textbf{32.35} & \textbf{59.02} & \textbf{82.72} & 86.31 & \textbf{76.24} & 89.45 & 91.47 \\
\bottomrule  
\end{tabular}  

\caption{All experiments use DeepSeek-R1-Distill-Qwen-1.5B as the base model. \textbf{Bold} indicates better performance between GRPO and MR-RLVR at the same data scale.} 
\end{table*}  

To investigate the data efficiency of MR-RLVR, we conduct experiments with different training data scales on DeepSeek-R1-Distill-Qwen-1.5B. Table~\ref{tab:data_efficiency} compares MR-RLVR against the GRPO baseline using 1k and 3k training samples.

Results show that MR-RLVR consistently outperforms GRPO across different data regimes. With 1k samples, MR-RLVR demonstrates significant improvements over GRPO, especially on Pass@5 and Pass@8 metrics, while Pass@1 shows minimal gains. On AIME24, MR-RLVR achieves 26.90\% and 31.84\% for Pass@5/Pass@8 compared to GRPO's 24.62\% and 28.69\%, representing relative gains of 9.26\% and 10.99\%, whereas Pass@1 remains unchanged at 11.09\%. This pattern suggests that with limited training data, the MR-RLVR framework primarily improves the model's ability to generate diverse high-quality candidates rather than directly enhancing single-sample accuracy. When scaled to 3k samples, MR-RLVR maintains its advantage with a 10.08\% relative improvement on AIME24 Pass@8 (40.97\% vs. 37.22\%). This consistent advantage suggests that process-level self-supervision in MR-RLVR provides more sample-efficient learning signals than standard GRPO, enabling better generalization with limited training data. On the simpler MATH500 benchmark, the gap narrows, confirming that MR-RLVR's benefits are most pronounced on challenging problems.

\subsection{MR-RLVR Tasks for Data Augmentation}

Given the sample efficiency gains demonstrated by MR-RLVR, we further explore self-supervised pretraining tasks for expanding training signals without additional human annotations. We design two tasks, step reordering and masked-then-fill, that leverage existing mathematical reasoning corpora to automatically generate diverse reasoning trajectories.  Table~\ref{tab:reordering_steps} and Table~\ref{tab:masked_then_fill} presents two representative cases together with model outputs.

\paragraph{Value of the step reordering task.} As shown in Table~\ref{tab:reordering_steps}, in the case involving the Lebesgue differentiation theorem, the model needs to restore 6 shuffled proof steps to their correct logical order. During this process, the model performs detailed logical analysis of each step and identifies inter-step dependencies. For instance, the model recognizes that Step 2 (defining \(F(x)\)) is the starting point of the proof, Step 4 (providing the dominating function) is a necessary condition for applying the Dominated Convergence Theorem, and Step 1 is the key theorem application step. This analytical process constitutes a structured reconstruction of the original proof: the model not only produces the correct ordering but also generates an explanation of why this ordering is valid. Compared to the original shuffled steps, the model automatically generates logical interpretations of each step and explicit annotations of inter-step dependencies during the reordering process. These generated reasoning trajectories effectively complement the original concise reasoning process. While original proofs typically only provide key steps, the model's analysis reveals how to identify the overall proof structure and how to determine logical dependencies between steps, thereby making implicit reasoning structures explicit. This automatically generated structured interpretation provides richer training signals for models to learn complete proof construction capabilities.

\paragraph{Value of the masked-then-fill task.} Table~\ref{tab:masked_then_fill} presents a masked-then-fill case involving bitwise operations. The task requires the model to complete three masked key formulas. The model successfully derives the first two: simplifying \(A \oplus \text{\texttt{0xFFFFFFFF}}\) to \(\sim A\), and determining \(A = \text{\texttt{0x81010100}}\) through bit analysis. However, at the third mask position (verification step), the model provides \(\text{\texttt{0x7EFEFEFF}} \oplus \text{\texttt{0x81010100}} = \text{\texttt{0xFFFFFFFF}}\), while the original solution requires computing the addition \(\text{\texttt{0x7EFEFEFF}} + \text{\texttt{0x81010100}}\). Although this equation is mathematically correct, since the XOR operation does yield \(\text{\texttt{0xFFFFFFFF}}\), it uses the wrong operator. More subtly, because these two numbers have no overlapping bits (no bit position is 1 in both numbers), addition and XOR happen to produce identical results in this case. This coincidentally correct situation reveals speculative behavior: the model may directly apply the same pattern after seeing XOR operators multiple times in the preceding text, or infer from context that the result should be \(\text{\texttt{0xFFFFFFFF}}\) and then reverse-engineer a seemingly reasonable formula, rather than strictly following the required reasoning procedure. Such errors are more difficult to detect than obvious computational mistakes.

Overall, the two tasks augment the training corpus in two complementary ways. First, correct trajectories generated during step reordering and masked-then-fill provide detailed and structured reasoning traces that can be directly reused as additional training data. Second, speculative errors surfaced by the **masked-then-fill task** can be turned into error-correction objectives, where models are trained to identify and fix logical flaws, thereby improving their self-checking capability. This pretraining stage therefore supplies more informative and sample-efficient learning signals than relying solely on supervised solutions.

\section{Conclusion}

This paper presents MR-RLVR, a framework that enriches reinforcement learning from verifiable rewards with process-level self-supervision. Instead of relying solely on outcome-level rewards derived from final-answer checking, MR-RLVR constructs two types of process-level tasks, namely masked-then-fill and step reordering, on mathematical reasoning traces, thereby providing dense training signals for intermediate reasoning steps. These tasks encourage the model not only to produce correct final answers, but also to acquire reusable patterns and structures of reasoning, rather than merely memorizing superficial solution templates.

We implement and evaluate MR-RLVR on Qwen2.5-3B and DeepSeek-R1-Distill-Qwen-1.5B, and conduct systematic experiments on a diverse set of mathematical benchmarks, including AIME24, AIME25, AMC23, and MATH500. Under a fixed sampling and decoding budget, MR-RLVR consistently outperforms standard RLVR. This indicates that process-level self-supervision becomes especially beneficial when problems require long-horizon, multi-step reasoning. Our data efficiency analysis further shows that, compared to relying solely on outcome-level rewards, MR-RLVR provides more informative learning signals in low-data regimes.

For future work, we first note that the current implementation adopts fixed masking positions for masked-then-fill task and a fixed shuffling scheme for step reordering. An interesting direction is to explore dynamically sampling masking locations and reordering strategies during training, allowing data augmentation and process-level tasks to adapt to the model's current state and further improve sample efficiency. Second, we plan to extend MR-RLVR to broader structured reasoning domains such as program synthesis and formal theorem proving, as well as to multimodal reasoning tasks involving images, diagrams, and geometric figures, where rich structure and verifiable signals naturally arise. In addition to masking and reordering, we aim to design more diverse process-level tasks, such as error correction tasks that explicitly require the model to identify and revise incorrect steps in a reasoning chain. Finally, MR-RLVR is highly complementary to explicit process reward models and test-time scaling techniques; integrating these components more tightly may further enhance the reliability and scalability of reasoning-focused language models. We hope that MR-RLVR offers a useful starting point for more principled integration of self-supervision and verifiable rewards in the training of reasoning-oriented large language models.

\bibliographystyle{plainnat} 
\bibliography{mr_rlvr}  

\clearpage
\appendix
\section*{Appendices}
\section{Prompts}
\label{app:prompts}
\subsection{Prompts for Data Curation} 
\begin{tcolorbox}[colback=brown!10!white, colframe=brown!75!black, boxrule=0.5mm, title=Prompt for masked-then-fill data curation]
You are a helpful assistant. 

Task: Extract the most key formulas or theorem names from the following original answer text and save them in JSON format.

Output Format: Return the key formulas or theorem names in a JSON object with the following structure:
\begin{verbatim}
{
    "theorems": [
        "Theorem or formula name 1",
        "Theorem or formula name 2",
        "Theorem or formula name 3",
        // Continue until all key formulas or theorem names are included
    ]
}
\end{verbatim}

\textbf{Requirements}:
\begin{itemize}
    \item Extract only the content from the original text without adding new formulas or theorems.
    \item Use standard LaTeX format for all mathematical symbols and expressions.
    \item Sort the extracted theorems by importance, placing the most important ones first and the less important ones later.
    \item The output must comply with JSON format and be ready for use.
\end{itemize}
\end{tcolorbox}

\begin{tcolorbox}[colback=brown!10!white, colframe=brown!75!black, boxrule=0.5mm, title=Prompt for step reordering data curation]
You are a helpful assistant. 

Task: Split the following answer into independent logical steps while maintaining the original meaning of the content.

Output Format: Return the steps in a JSON object with the following structure:
\begin{verbatim}
{
    "steps": [
        "Step 1 description...",
        "Step 2 description...",
        "Step 3 description...",
        // Continue until all steps are included
    ]
}
\end{verbatim}

\textbf{Requirements}:
\begin{itemize}
    \item All steps must be generated from the original answer text without creating new steps or content.
    \item Each step should maintain an independent logical meaning, allowing it to stand alone.
    \item The steps should connect logically in a way that reconstructs the original answer when combined together.
    \item Ensure clarity and conciseness in each step to facilitate understanding.
    \item Use standard LaTeX format for all mathematical symbols and expressions.
\end{itemize}
\end{tcolorbox}
\subsection{Prompts for MR-RLVR}
\textcolor{white}{\textbf{blank}}
\begin{tcolorbox}[colback=brown!10!white, colframe=brown!75!black, title=Prompt for Masked-Then-Fill task]
\texttt{System: } \\
A conversation between the User and the Assistant.

The User supplies a mathematical statement together with a partial solution in which some formulas or theorems are masked with \texttt{<formula\_masked>} tags.

The Assistant's task is to complete the missing portions of the solution by replacing the \texttt{<formula\_masked>} tags with the appropriate mathematical formulas or theorems.

Please adhere to the following structured approach:
\begin{quote}
1. Begin by performing a comprehensive logical analysis to determine the precise formula required for each \texttt{<formula\_masked>} tag. The objective is to ensure the logical coherence and completeness of the entire solution.

2. Enclose your **detailed logical analysis**, explaining the derivation of each missing formula, within \texttt{<think>} tags, formatted as follows:
\begin{quote}
\texttt{<think>} \\
\hspace{1em} [Your detailed reasoning process, explaining how each missing formula was derived.] \\ 
\texttt{</think>}
\end{quote}

3. Finally, upon completion of the analysis and derivation of all missing formulas, provide **only** the derived formulas, enclosed within \texttt{\textbackslash boxed\{\}} notation:
\begin{quote}
\texttt{\textbackslash boxed\{formula\_1; formula\_2; ...; formula\_n\}}
\end{quote}
The formulas within the \texttt{\textbackslash boxed\{\}} answer must appear in the same order as their corresponding \texttt{<formula\_masked>} tags in the original solution. All mathematical formulas should be presented using proper LaTeX notation.
\end{quote}
\texttt{User: } \\
The user's statement: \\
The partial solution is: 
\end{tcolorbox}

\begin{tcolorbox}[colback=brown!10!white, colframe=brown!75!black, title=Prompt for Step reordering task]
\texttt{System: } \\
A conversation between the User and the Assistant.

The User supplies a mathematical statement and a solution whose steps are out of order (each step is already numbered with 'Step i').

The Assistant's task is to determine the correct logical sequence of these steps.

Please adhere to the following structured approach:

\begin{enumerate}
    \item Begin by performing a comprehensive logical analysis of the mathematical statement and all given steps to establish their correct sequential order. The objective is to reconstruct a logically sound and complete solution.

    \item Enclose your **detailed logical analysis**, explaining how you determined the correct sequence, within \texttt{<think>} tags, formatted as follows:
    \begin{quote}
    \texttt{<think>} \\
    \hspace{1em} [Your detailed reasoning process, explaining how the logical sequence of steps was determined.] \\ 
    \texttt{</think>}
    \end{quote}

    \item Finally, provide **only** the correct sequence of step numbers, enclosed within \texttt{\textbackslash boxed\{\}} notation:
    \begin{quote}
    \texttt{\textbackslash boxed\{n1, n2, n3, \dots, nk\}}
    \end{quote}
    The step numbers within the \texttt{\textbackslash boxed\{\}} answer must represent the final, logically ordered sequence of the steps.
\end{enumerate}
\texttt{User: } \\
The user's statement: \\
The shuffled solution:
\end{tcolorbox}

\begin{tcolorbox}[colback=brown!10!white, colframe=brown!75!black, title=Prompt for Outcome-only Task]
\texttt{System: } \\
A conversation between User and Assistant.

The User provides a question, and the Assistant outputs the answer.

The Assistant’s task is to solve the question and provide the final answer.

Please adhere to the following structured approach:

\begin{enumerate}
    \item Provide a concise solution analysis to determine how to compute the answer and enclose a detailed, step-by-step derivation within \texttt{<think>} tags. Use the following format:
    \begin{quote}
    \texttt{<think>} \\
    \hspace{1em} [Your detailed reasoning process analysis, explained through a step-by-step derivation.] \\ 
    \texttt{</think>}
    \end{quote}

    \item Finally, provide only the final result written in standard LaTeX and enclosed within \texttt{\textbackslash boxed\{ \}}.
\end{enumerate}
\texttt{User: } \\
he user's question: 
\end{tcolorbox}

\section{Implementation Details}
\label{app:impl}
\begin{table}[ht]
    \centering
    \begin{tabular}{|l|c|c|}
        \hline
        \textbf{Parameter} & \textbf{qwen-3b STAGE I} & \textbf{qwen-3b STAGE II} \\ 
        \hline
        Learning Rate ($\text{lr}$) & $1 \times 10^{-6}$ & $1 \times 10^{-6}$ \\ 
        \hline
        Rollout Number  & 16 & 16 \\ 
        \hline
        Rollout Temperature  & 1.0 & 1.0 \\ 
        \hline
        Prompt Length Token  & 2048 & 1024 \\ 
        \hline
        Response Length Token  & 4096 & 4096 \\ 
        \hline
        Training Batch Size  & 512 & 512 \\ 
        \hline
        PPO Mini Batch Size & 64 & 64 \\ 
        \hline
        KL Loss Coefficient  & 0.001 & 0.001 \\ 
        \hline
        Training Epochs  & 3 & 3 \\ 
        \hline
    \end{tabular}
    \caption{The training hyperparameters of MR-RLVR for qwen-3b}
    \label{tab:qwen_3b_params}
\end{table}

\begin{table}[ht]
    \centering
    \begin{tabular}{|l|c|c|}
        \hline
        \textbf{Parameter} & \textbf{Deepseek-R1-distill-Q
        wen-1.5b STAGE I} & \textbf{Deepseek-R1-distill-Q
        wen-1.5b STAGE II} \\ 
        \hline
        Learning Rate & $1 \times 10^{-6}$ & $1 \times 10^{-6}$ \\ 
        \hline
        Rollout Number & 8 & 16 \\ 
        \hline
        Temperature  & 1.0 & 1.0 \\ 
        \hline
        Prompt Length Token  & 2048 & 1024 \\ 
        \hline
        Response Length Token  & 8192 & 4096 \\ 
        \hline
        Training Batch Size  & 512 & 512 \\ 
        \hline
        PPO Mini Batch Size  & 64 & 64 \\ 
        \hline
        KL Loss Coefficient  & 0.001 & 0.001 \\ 
        \hline
        Training Epochs  & 3 & 3 \\ 
        \hline
    \end{tabular}
    \caption{The training hyperparameters of MR-RLVR for deepseek-dis-qwen-1.5b}
    \label{tab:deepseek_dis_qwen_params}
\end{table}
\section{Test Cases}
\label{appendix:test_cases}
\textcolor{white}{\textbf{blank}}
\begin{table}[!htbp]
\centering
\begin{tcolorbox}[colback=blue!10!white, colframe=blue!75!black, sharp corners=south, boxrule=2pt, title={Case of Reordering Steps Task}]
\begin{flushleft} 
\textbf{Problem:} Leibniz Integral Rule under Lebesgue Conditions

Let \( f: \mathbb{R}^n \to \mathbb{R} \) satisfy the conditions of the Lebesgue differentiation theorem, and let \( S \subset \mathbb{R}^n \) be measurable with \( 0 < |S| < \infty \). 

Prove: For all \( x \) such that \( \frac{\partial f}{\partial x} \) exists a.e. in \( S \),

\[
\frac{d}{dx} \int_S f(x,y) \, dy = \int_S \frac{\partial f}{\partial x}(x,y) \, dy.
\]

\textbf{Input:} \\

Step 0: State that for almost every \( y \in S \), the pointwise limit 
\[
\lim_{h \to 0} \frac{f(x+h,y) - f(x,y)}{h} = \frac{\partial f}{\partial x}(x,y)
\]
exists by hypothesis. \\

Step 1: Apply the Dominated Convergence Theorem to interchange limit and integral, yielding 
\[
\lim_{h \to 0} \int_S \frac{f(x+h,y) - f(x,y)}{h} \, dy = \int_S \frac{\partial f}{\partial x}(x,y) \, dy.
\] \\

Step 2: Define \( F(x) = \int_S f(x,y) \, dy \) and state the goal to show 
\[
\frac{dF}{dx} = \int_S \frac{\partial f}{\partial x}(x,y) \, dy.
\] \\

Step 3: Conclude that 
\[
\frac{dF}{dx} = \int_S \frac{\partial f}{\partial x}(x,y) \, dy.
\] \\

Step 4: Note that \( \left| \frac{\partial f}{\partial x}(x,y) \right| \leq g(y) \) and 
\[
\left| \frac{f(x+h,y) - f(x,y)}{h} \right| \leq g(y) 
\]
for sufficiently small \( h \), where \( g \) is Lebesgue integrable over \( S \). \\

Step 5: Express the difference quotient for \( F(x) \) as 
\[
\frac{F(x+h) - F(x)}{h} = \int_S \frac{f(x+h,y) - f(x,y)}{h} \, dy \quad \text{for } h \neq 0.
\]

\textbf{Output:} \\

Step 0: This is a crucial step in establishing the differentiability of \( f \) in the \( x \)-direction. \\

Step 1: This is a key step in proving the differentiability of \( F(x) \). \\

Step 2: This is the starting point of the proof. \\

Step 3: This is the final result we are aiming to prove. \\

Step 4: This is a necessary step to apply the Dominated Convergence Theorem. \\

Step 5: This is a crucial step in establishing the differentiability of \( F(x) \). \\

Based on this analysis, the correct logical sequence is:

\[
2 \to 5 \to 0 \to 4 \to 1 \to 3
\]

\end{flushleft}
\end{tcolorbox}
\caption{Detailed Examination of Reordering Steps in the Proof of the Leibniz Integral Rule under Lebesgue Conditions}
\label{tab:reordering_steps}
\end{table}

\begin{table}[!htbp]
\centering
\begin{tcolorbox}[colback=blue!10!white, colframe=blue!75!black,  boxrule=2pt, title={Case of Masked-Then-Fill Task}, sharp corners=south]

\begin{flushleft} 
\textbf{Problem:}  Find the hexadecimal value of \( A \) satisfying 

\[
(\texttt{0x7EFEFEFF} + A) \oplus (A \oplus \texttt{0xFFFFFFFF}) = \texttt{0x81010100}
\]

where \( + \) denotes bitwise addition and \( \oplus \) denotes XOR.
\\ 
\textbf{Input:} \\

Step 1. Simplification \\
Note that \( A \oplus \texttt{0xFFFFFFFF} = \sim A \) for any 32-bit value. Rewrite the equation as: 

\[
\text{<formula\_masked>}
\]

Step 2. Binary Analysis \\
Write \( \texttt{0x7EFEFEFF} \) in binary: 

\[
\texttt{0x7EFEFEFF} = 01111110 \, 11111110 \, 11111110 \, 11111111 
\]

Observe: zeros at bit positions 31, 24, 16, and 8.

Step 3. Deriving A \\
To avoid carries during addition, \( A \) should have 1's only where \( \texttt{0x7EFEFEFF} \) has 0’s: 

\[
\text{<formula\_masked>}
\]

Step 4. Verification \\
Compute \( \texttt{0x7EFEFEFF} + A \). Since there is no overlap (no bit is 1 in both numbers), the addition result is:

\[
\text{<formula\_masked>}
\]

Then verify: 

\[
\texttt{0xFFFFFFFF} \oplus \texttt{0x7EFEFEFF} = \texttt{0x81010100} 
\]

\textbf{Output:} \\
\[
\text{<formula\_masked>} \rightarrow (\texttt{0x7EFEFEFF} + A) \oplus \sim A = \texttt{0x81010100} \quad \checkmark \text{ Correct}
\]

\[
\text{<formula\_masked>} \rightarrow A = \texttt{0x81010100} \quad \checkmark \text{ Correct}
\]

\[
\text{<formula\_masked>} \rightarrow \texttt{0x7EFEFEFF} \oplus \texttt{0x81010100} = \texttt{0xFFFFFFFF} \quad \times \text{ Wrong operator! Should be '+'}
\]

\text{Expected: } \( \texttt{0x7EFEFEFF} + \texttt{0x81010100} = \texttt{0xFFFFFFFF} \)
\end{flushleft}
\end{tcolorbox}
\caption{Detailed Examination of Mask-Then-Fill for Finding the Hexadecimal Value of \( A \) Based on Bitwise Operations}
\label{tab:masked_then_fill}
\end{table}

\end{document}